\documentclass{article}

\usepackage{hyperref}
    \hypersetup{
        colorlinks,
        citecolor=blue,
        filecolor=blue,
        linkcolor=blue,
        urlcolor=blue
    }

\usepackage{booktabs}
\usepackage{multirow}
\usepackage{pgfplots}
    \pgfplotsset{compat=newest}
\usepackage{amssymb}
\usepackage{amsmath}
\usepackage{bm}
\usepackage[ruled,linesnumbered,lined,boxed]{algorithm2e}
\usepackage{tikz}
    \usetikzlibrary{positioning}
    \usetikzlibrary{shapes,arrows}
    \usetikzlibrary{shapes.geometric}
    \usetikzlibrary{shadows}
    \usetikzlibrary{calc}
\usepackage{graphicx}

\graphicspath{{./figs/}}

\DeclareMathOperator*{\argmax}{arg\,max}

\newcounter{x}

\providecommand{\keywords}[1]
{
  \small	
  \textbf{\textit{Keywords---}} #1
}

\begin{document}

\title{Multi-Task Handwritten Document Layout Analysis}

\author{Lorenzo Quir\'os}
\date{loquidia@prhlt.upv.es\\%
    PRHLT Research Center, 
    Universitat Polit\`ecnica de Val\`encia,
    Valencia, Spain}
\maketitle
\begin{abstract}
Document Layout Analysis is a fundamental step in Handwritten Text Processing
systems, from the extraction of the text lines to the type of zone it
belongs to. We present a system based on artificial neural networks
which is able to determine not only the baselines of text lines present in the
document, but also performs
\emph{geometric and logic layout analysis} of the document.
Experiments in three different datasets demonstrate the potential of
the method and show competitive results with respect to state-of-the-art methods. 
\end{abstract}



\section{Introduction}\label{sec:introduction}

Handwritten Text Processing (HTP) systems such as Handwritten 
Text Recognition (HTR)~\cite{romero2011}, Keyword 
Spotting (KWS)~\cite{Bluche2017} and Information Retrieval from Handwritten 
Documents~\cite{fornes2017} are well-known problems where an image of a 
handwritten document is used as an
input and some kind of text-related information is expected as output. But for 
all current HTP systems, the image is expected to contain just a single short 
handwritten sequence; that is, only 
one \emph{line} of handwritten text is processed at once. However, since the 
main goal of those systems is to process not just a single line, but a
complete paragraph or even a complete document, a previous system is needed in
order to extract the required lines from the whole page and, in an upper level, to 
segment the different zones of the page (paragraph, marginal notes, 
illustrations, page number, etc.) in a meaningful manner (normally consistent 
with the reading order). 

Consequently, both test line extraction and image segmentation into relevant 
zones constitute a very important stage of any HTP system, generally related as
Document Layout Analysis (DLA).
Commonly this process is 
divided into two sub problems~\cite{cattoni1998}. First, \emph{geometric layout analysis}
aims at producing a description of the geometric structure of the document ( i.e.
where each zone is placed, its shape and relationship with other zones). This 
structure allows us to describe the document layout at different levels of detail 
(e.g. a set of text lines can be viewed at higher level as a paragraph). Second,
the classification of those zones into their logical role (title, paragraph, 
illustration, etc.) is called the \emph{logical layout analysis}.  
Although, \emph{logical layout analysis} is not necessary for zone segmentation 
nor for text line
extraction, it is a very important step required to present the results of HTP 
systems in the same context as the input document (e.g. the transcript of some 
text line can be presented as a title or other type only if the zone label is
defined, otherwise we can just provide the plain transcript).


In most recent formulations the determination of text lines focus on the detection of
their baselines (the imaginary lines upon which the lines of text rest) rather than a
detailed polygons surrounding the text lines.  Owing to the fact that a baseline is defined by 
only a few points, humans can easily label on the text lines of an image, without
having to deal with the cumbersome detailed segmentation of each text line region.

Once some DLA system provides a
baseline for each text line in a document it can be easily reviewed and corrected
by the user. 
Also, rough segmentation of a text line can be straightforwardly obtained from
its baseline, and because state-of-the-art HTP systems are able to filter out a 
lot of noise present in the input, this roughly segmented lines can be used by the HTP
system with almost no negative impact on performance~\cite{Romero2015}.


It is very important to notice the huge impact the context provided by the
\emph{logical layout analysis} can have in the 
performance of HTP systems. For example a well segmented text line labeled as
a part of the \emph{page number} text zone is expected to have only digits; then 
the search space for the HTP system can be reduced drastically.

In this work, we present a system based on Artificial Neural Networks, which is 
able to detect the baselines, layout zones, and labels of that zones, from the 
digital image of the document.
It is an integrated approach where baselines and text zones
are detected and segmented in a single process, and the relationship between them is 
defined in a top-down way.

The rest of this paper is organized as follows. In Sec.~\ref{sec:RelWork}
related work is discussed. Then, the proposed method is presented in 
Sec.~\ref{sec:method}. Afterwards, experimental setup is presented in 
Sec.~\ref{sec:exp}, while results are reported in Sec.~\ref{sec:results}.
Finally, we draw some conclusions in Sec.~\ref{sec:conclusions}.

\section{Related work}
\label{sec:RelWork}
Comprehensive surveys about document image analysis~\cite{Nagy2000,Mao2003,Namboodiri2007}
and \cite{Eskenazi2017} provide a very good 
insight about the state-of-the-art algorithms for document segmentation,
including DLA.

DLA methods can be divided typologically into three groups by the problem they 
are developed to solve: text line 
extraction (included baseline detection), zone segmentation and zone labeling.
Most of the methods
focus on only one of these groups, or provide a separate algorithm
for each one. In contrast, the method we present in this work encompasses all 
three groups under the same model.

\subsection{Text line extraction}
This is the group to which most methods belong to, mainly because its direct 
applicability to HTP systems. The main goal of these methods is to segment the 
input image into a set of smaller images, each of which contains a single text 
line.

Methods such as those presented in~\cite{Shi2009, Ryu2014, Ouwayed2012, Cohen2014, Baechler2013}
rely on connected components extraction after some filtering stage, while other 
methods such as 
\cite{Arvanitopoulos2014, Nicolaou2009} use a tracer function to separate 
the different text lines after applying a blur to the input image. Other methods
rely on Hidden Markov Models \cite{Bosch2012,Bosch2012_b,Bosch2014} or 
Recurrent Neural Networks \cite{Moysset2015} to model the vertical sequential structure
of the text lines, or Convolutional Neural Networks \cite{Pastor2016,Gruning2018,Ares2018} to classify 
each pixel of the image between text line and non-text line.


\subsection{Zone segmentation}
Most of the methods for text line extraction rely on the assumption that input
images contain a single region of text; which means, documents with a single 
column layout, or images previously segmented into the different text zones. 
Zone segmentation
aims at providing this level of page image segmentation.

Several methods are based on some kind of pixel-level classifier (Multilayer Perceptron 
\cite{Bukhari2012, Baechler2011, Wei2013, Baechler2013}, 
Conditional Random Fields \cite{Cruz2012}, 
Definite Clause Grammars \cite{Lemaitre2008},
Gaussian Mixture Models and Support Vector Machines \cite{Wei2013})
whose input is a set of handcrafted features from
the input images (Gabor filters, Kalman filters, Connected Components, 
Multi-scale images, etc).
Others aim to provide an interactive framework to review the results
of the zone segmentation algorithm \cite{Quiros2017}.

\subsection{Zone labeling}


Methods of this group are often closely related with methods of the previous one,
but some of 
them focus only on separating text from non-text zones \cite{Zhong2015, Wei2013,Baechler2011,Ares2018} 
(which can be considered as a simplified form of Zone labeling).
Other approaches go further to provide not just the segmentation of the zones
but also the corresponding zone labels (three different zones are labeled in \cite{Cruz2012},
two in \cite{Bukhari2012} and six in \cite{Lemaitre2008}).

Any of the three groups listed before can be used to help the processing of any other
group (e.g. segmented zones can be used to constrain the search space for line detection
and vice-versa), which is a causality dilemma in the design of any DLA system.
%
An integrated method, as proposed in this paper, provides a solution to this 
dilemma where the relevant
dependencies are incorporated internally in the model.

\section{Proposed Method}\label{sec:method}

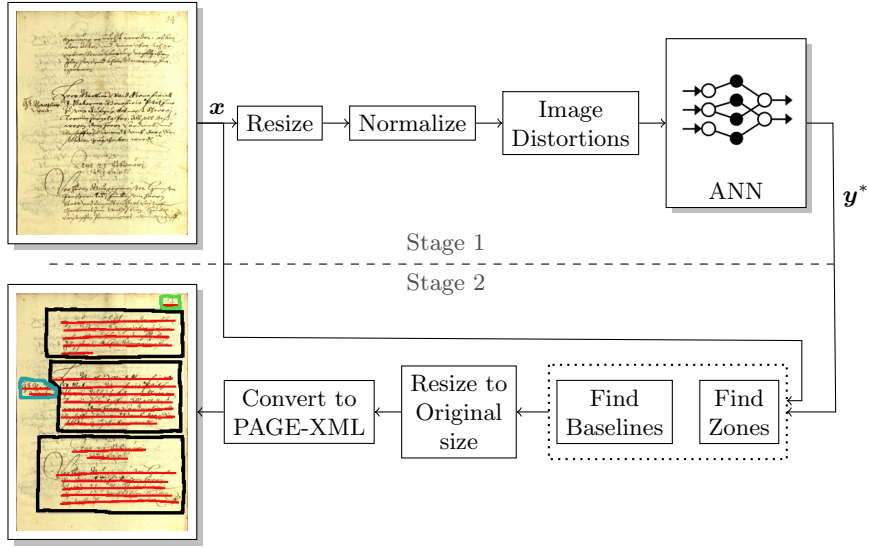
\begin{figure*}
    \centering
    \begin{tikzpicture}[font=\small]
        \node[rectangle, draw, fill=white,drop shadow] (0,0) (IN) {\includegraphics[width=7em]{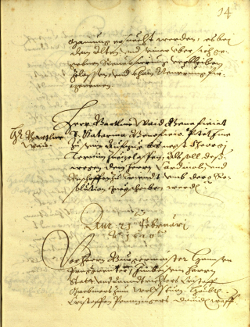}};
        \node[rectangle, draw, right=1.5em of IN] (0,0) (RES) {Resize};
        \node[rectangle, draw, right=1em of RES] (0,0) (NORM) {Normalize};
        \node[rectangle, draw, right=1em of NORM,align=center] (0,0) (DIST) {Image\\Distortions};
        \node[rectangle, draw, right=1em of DIST, fill=white,drop shadow,align=center] (0,0) (ANN) {\includegraphics[width=5em]{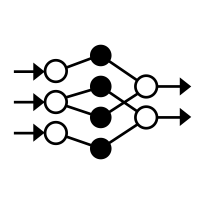}\\ANN};
        \node[rectangle, draw, below=1.5em of IN,fill=white, drop shadow] (0,0) (OUT) {\includegraphics[width=7em]{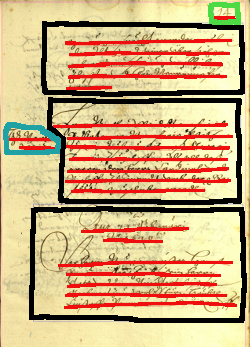}};
        \node[rectangle, draw, right=1em of OUT,align=center] (0,0) (CONV) {Convert to\\PAGE-XML};
        \node[rectangle, draw, right=1em of CONV,align=center] (0,0) (nRES) {Resize to\\Original\\size};
        \node[rectangle, draw, right=1.5em of nRES,align=center] (0,0) (FR) {Find\\Baselines};
        \node[rectangle, draw, right=1em of FR, align=center] (0,0) (FB) {Find\\Zones};
        \draw[thick,dotted] ($(FR.north west)+(-0.1,0.2)$) rectangle ($(FB.south east)+(0.1,-0.2)$);

        \draw[->] (IN) -- node[above] {$\bm{x}$} (RES);
        \draw[->] (RES) -- (NORM);
        \draw[->] (NORM) -- (DIST);
        \draw[->] (DIST) -- (ANN);
        \path[->,draw] (ANN) -- ($(ANN.east)+(0.35,0.0)$) -- node[auto,near start]{$\bm{y}^{*}$} ($(FB.east)+(0.75,0.0)$) -- ($(FB.east)+(0.1,0.0)$);
        \path[->,draw] ($(FR.west)+(-0.1,0.0)$) -- (nRES);
        \draw[->] (nRES) -- (CONV);
        \draw[->] (CONV) -- (OUT);
        \draw[->] (IN) -- ($(IN.east)+(0.35,0.0)$) -- ($(IN.east)+(0.35,-2.85)$) -- ($(FB.east)+(0.3,0.95)$) -- ($(FB.east)+(0.3,0.15)$) -- ($(FB.east)+(0.1,0.15)$);

        \path[draw,dashed,black!70] ($(IN.south)+(-2em,-0.75em)$) -- node[midway,auto]{Stage 1} ($(IN.south)+(28em,-0.75em)$);
        \path[draw,dashed,black!70] ($(IN.south)+(-2em,-0.75em)$) -- node[midway,auto,swap]{Stage 2} ($(IN.south)+(28em,-0.75em)$);

    \end{tikzpicture}
    \caption{Proposed method overview. $\bm{x}$ is the input image and 
    $\bm{y}^{*}$ is the best hypothesis from the ANN. Better seeing in color.}
    \label{fig:P2PaLA}
\end{figure*}

An overview of the proposed method for Document Layout Analysis is given in 
Fig.~\ref{fig:P2PaLA}. The method consists of two main stages.
\begin{itemize}
    \item Stage 1: Pixel level classification, for zones and baselines.
    \item Stage 2: Zone segmentation and baseline detection.
\end{itemize}
In the first 
stage an Artificial Neural Network (ANN) is used to classify the pixels of the 
input image ($\bm{x} \in \mathbb{R}^{w\times h \times \gamma}$, with height $h$,
width $w$ and $\gamma$ channels)
into a set of regions of interest (layout zones and baselines). This is a crucial 
stage in the process, where information from the images is extracted, while next stage 
is designed to provide that information in a usefull format.

In the second state a contour extraction algorithm is used to consolidate the pixel 
level classification into a set of simplified zones delimited by closed 
polygons. Then a similar process is carried on inside each zone to extract 
the baselines. In this way we obtain the location and label of each zone, and
the baselines that they contain.



\subsection{Stage 1: Pixel Level Classification}
\label{sec:stage1}
Layout Analysis can be defined as a Multi-Task 
problem~\cite{Caruana93multitasklearning:} where two tasks are defined:

\begin{itemize}
\item \emph{Task-1}: Baseline detection.
\item \emph{Task-2}: Zone segmentation and labeling.
\end{itemize}

\emph{Task-1} consists in obtaining the baseline of each text line 
present in the input image. This baseline is used to extract the sub-image of the
text line and feed it into some HTP system.
On the 
other hand, \emph{Task-2} consists in assigning those baselines into the
different zones they belong to. For example, baselines that are 
members of the main paragraph should be grouped together into a zone, while the 
ones that are members of a marginal note should be
grouped into another one. 
Since each line belongs to a different context, that information can be used by 
the HTP system to provide more accurate hypotheses.

In a general manner we can define a multi-task variable\footnote{For convenience,
each task will be represented mathematically as a superscript over the variables (e.g. $v^t$).}
$\bm{y}=[\bm{y}^1,\dots,\bm{y}^T]$,
where $\bm{y}^t=(y^{t}_{ij}), 1 \le i \le w, 1 \le j \le h, 1 \le t \le T$ and $y^{t}_{ij} \in \mathcal{Y}^t = \{1,\dots,K^t\}^{w\times h}$
with $K^t \in \mathbb{N}_+$ 
being the finite number of classes associated with the $t$-th task.
The solution of this problem for some test instance $\bm{x}$ is given as the
following optimization problem:

\begin{equation}
	\label{eq:class}
    \hat{\bm{y}} = \argmax_{\bm{y}} \mathrm{p}(\bm{y}\mid\bm{x})
\end{equation}
where the conditional distribution $\mathrm{p}(\bm{y}\mid\bm{x})$ is usually 
unknown and has to be estimated from training data $D=\{(\bm{x}_i,\bm{y}_i)\}^{N}_{n=1}=\{(\bm{X},\bm{Y})\}$.

In our specific two-task case ($T=2$), \emph{Task-1} ($t=1$) is a binary 
classification problem, then $K^1=2$ (background, baseline). On the other hand, 
\emph{Task-2} ($t=2$) is a multi-class problem where $K^2$ is equal to the 
number of different types of zones in the specific corpus, plus one for the 
background; normally, the number of types of zones is small (e.g. $K^{2} < 15$).

\begin{figure}[!t]
	\centering
    \includegraphics[width=\linewidth]{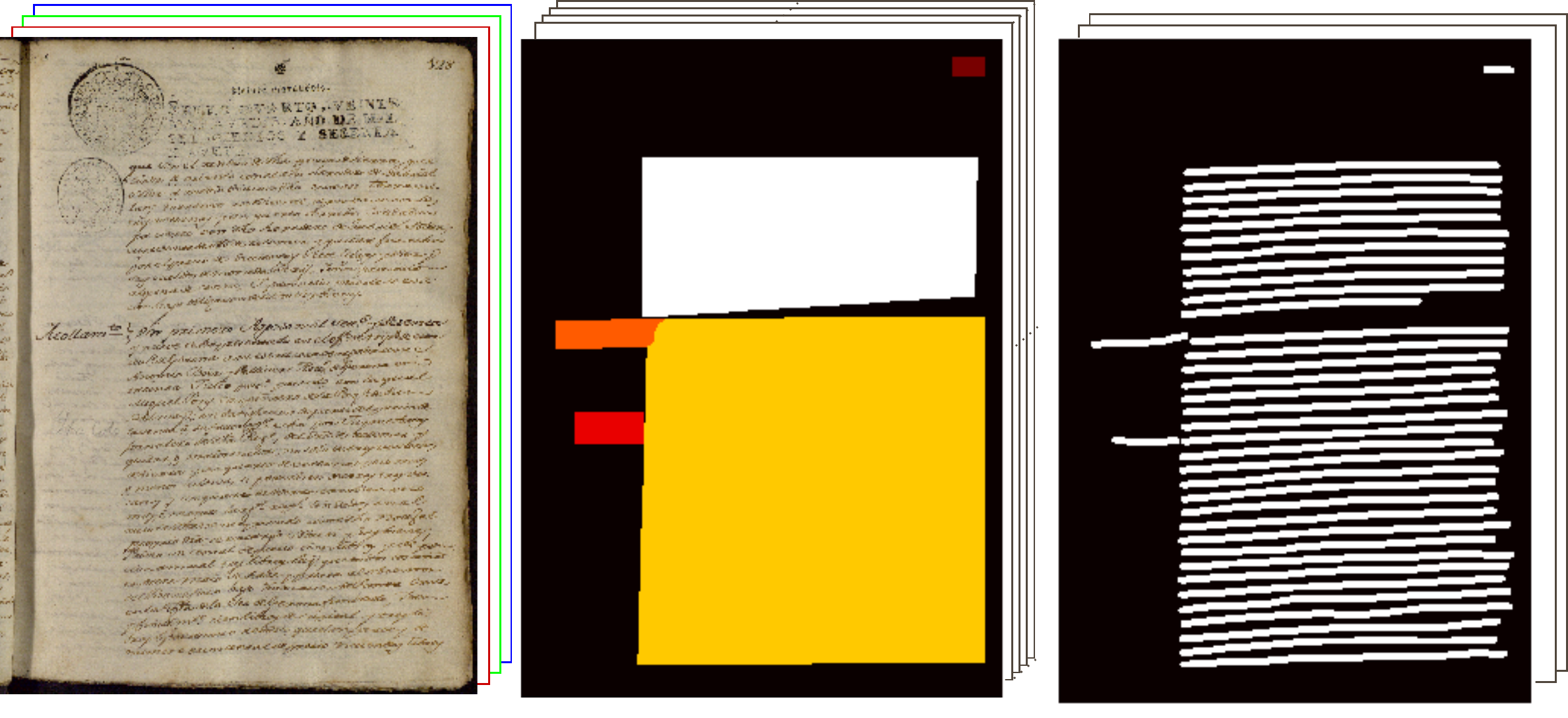}
    \caption{Visualization of the encoded ground-truth to feed the ANN during 
    training. From left to right: Original 
    image ($\bm{x}$), \emph{Task-2} pixel-level ground-truth ($\bm{y}^{t=2}$), \emph{Task-A} pixel-level 
    ground-truth ($\bm{y^{t=1}}$). The colors in the ground-truth images represents
    the class label of each task. Better seeing in color.}
    \label{fig:label}
\end{figure}

The Conditional Generative Adversarial Network presented in~\cite{pix2pix2016} has
shown very good results in several problems versus the non-adversarial neural networks.
In this work the conditional distribution $\mathrm{p}(\bm{y}\mid\bm{x})$ is 
estimated by a modified version of the Conditional Generative Adversarial Network,
where the  output layer of the generative network is replaced by a 
softmax layer for each task to ground the network to the underlying
discriminative problem (this can be considered as a type of 
Discriminative Adversarial Network, as presented in~\cite{SantosWZ17}). The
ANN is trained using labeled data as depicted in Fig.~\ref{fig:label}, where the
color represents the class label of each task.

\subsubsection{ANN Inference}
\begin{figure}
	\centering
    \tikzstyle{block} = [draw, fill=white, rectangle, minimum height=3em, minimum width=3em, drop shadow]
    \tikzstyle{mux} = [draw,trapezium, shape border uses incircle, shape border rotate=270]
    \tikzstyle{input} = [cylinder,
        cylinder uses custom fill,
        shape border rotate=90,
        aspect=0.25,
        minimum height=3em,
        draw]
    \begin{tikzpicture}[auto,>=latex']
        \node[cylinder, shape border rotate=90, aspect=0.25,minimum height=3em,draw] (IN) at (0,0) {$D$};
        \node[block,draw, right of=IN,node distance=4em] (M) {\emph{M-net}};
        \node[rectangle,draw, right= 5em of M] (arg) {\small{$\displaystyle \argmax_{\bm{y}}(\cdot)$}};
        \node[ trapezium, rotate=90,trapezium stretches body, shape border rotate=180,above right = -1.2em and 3em of arg] (MUX) {};
 
        \draw[->] (IN) -- node {$\bm{x}$} (M);
        \draw[->] (M) -- node {$\mathcal{M}(\bm{x})$} (arg);
        \draw[->] (arg) -- node[swap] {$\bm{y^{*}}$} (MUX.north west);
    \end{tikzpicture}
    \caption{Inference set-up for ANN.}
    \label{fig:DAN_inf}
\end{figure}
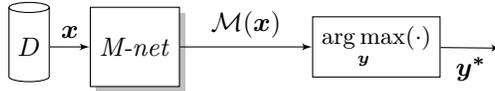

An ANN, called \emph{M-net}, is trained, as discussed in Sec.~\ref{sec:objective},
to estimate the posterior probability in 
Eq.~(\ref{eq:class}), as depicted in Fig.~(\ref{fig:DAN_inf}) 
%
where $\mathcal{M}(\cdot)$ is the output of the latest layer of \emph{M-net}, and 
the $\argmax$  operation is computed element-wise and separately for each task 
involved, this is:
\begin{itemize}
    \item \emph{Task-1}:
        \begin{equation}
            y^{*1}_{i,j} = \argmax_{y\in \mathcal{Y}^1 = \{0,1\}} \mathcal{M}_{i,j,y}(\bm{x}),\quad 0\le i \le w, 0\le j \le h
        \end{equation}
    \item \emph{Task-2}:
        \begin{equation}
            y^{*2}_{i,j} = \argmax_{y\in \mathcal{Y}^2 = \{1,\dots,K^2\}} \mathcal{M}_{i,j,y}(\bm{x}) ,\quad 0\le i \le w, 0\le j \le h
        \end{equation}
\end{itemize}

Notice this optimization problem rather corresponds to a simplified model, where no
restrictions are formulated based on the prior knowledge we have of the problem
(e.g. a page number zone is not expected to be between paragraphs). Although some
prior knowledge is learned by the ANN during training, the experience 
in other areas such as HTR and KWS has demonstrated very positive results when
priors are explicitly considered\cite{Bluche2017}.
Consequently, in a future version of the proposed method a set of structural
restrictions, modeled as a prior-probability of $\bm{y}$, will be added to
take into account that valuable knowledge.


\subsubsection{ANN Objective and Training}\label{sec:objective}
Training an ANN is very dependent on the selected objective function, since it 
defines what we want to learn. Classical objective functions are defined 
directly by how we want to approximate the probabilistic distribution of the 
data (mean square error, cross entropy, etc), while new adversarial networks 
are using a composed objective function to improve performance. 

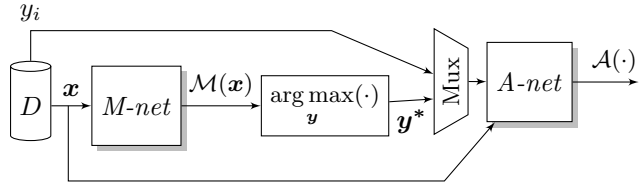
\begin{figure}
	\centering
    \tikzstyle{block} = [draw, fill=white, rectangle, minimum height=3em, minimum width=3em, drop shadow]
    \tikzstyle{mux} = [draw,trapezium, shape border uses incircle, shape border rotate=270]
    \tikzstyle{input} = [cylinder,
        cylinder uses custom fill,
        shape border rotate=90,
        aspect=0.25,
        minimum height=3em,
        draw]
    \begin{tikzpicture}[auto,>=latex']
        \node[cylinder, shape border rotate=90, aspect=0.25,minimum height=3em,draw] (IN) at (0,0) {$D$};
        \node[block,draw, right of=IN,node distance=4em] (M) {\emph{M-net}};
        \node[rectangle,draw, right= 3em of M] (arg) {\small{$\displaystyle \argmax_{\bm{y}}(\cdot)$}};
        \node[draw, trapezium, rotate=90,trapezium stretches body, shape border rotate=180,above right = -0.9em and 3em of arg] (MUX) {\small{Mux}};
        \node[block,above right = -2.2em and 2em of MUX] (A) {\emph{A-net}};
 
        \draw[->] (IN) -- node {$\bm{x}$} (M);
        \draw[->] (M) -- node {\small{$\mathcal{M}(\bm{x})$}} (arg);
        \draw[->] (arg) -- node[swap,midway, below] {$\bm{y^{*}}$} (MUX.north west);
        \draw[->] (MUX.top side) -- node {} (A);
        \draw[->] (IN) |- node {$y_i$} (4,1) -- (MUX);
        \draw[->] (0.5,0) |- (0.5,-1) -- (5.5,-1) -- (A) ;
        \draw[->] (A) -- node {~~\small{$\mathcal{A}(\cdot)$}} (23em,0.9em) ;
    \end{tikzpicture}
    \caption{Training set-up for ANN.}
    \label{fig:DAN}
\end{figure}

The objective function we want to minimize is composed of the interaction between
two separated ANNs, that we call 
\emph{A-net} and \emph{M-net}, see Fig.~\ref{fig:DAN}. The \emph{A-net} is trained to distinguish 
between real labels (from $D$) and  produced labels (sometimes called fake 
labels) from \emph{M-net} (notice \emph{A-net} is used only to help train
\emph{M-net}, and is discarded at inference time). 

In the \emph{A-net} network the cost function is 
the classical cross-entropy loss,
where only two classes are defined, ``1'' when the input of the network belongs to 
the real labels, and ``0'' when the labels are generated by \emph{M-net}, that
is:\footnote{Parameters of the network 
are not shown explicitly in order to keep the notation as simple as possible.}
\begin{equation}
    \mathcal{L}_A(\bm{X},\bm{Y}) = \frac{1}{2} \left\{\mathcal{L}_{A}^{1}(\bm{X},\bm{Y}) +
    \mathcal{L}_{A}^{0}(\bm{X},\bm{Y})\right\}, \quad(\bm{X},\bm{Y}) \in D
\end{equation}
according to
\begin{equation}
    \mathcal{L}_{A}^{1}(\bm{X},\bm{Y}) = \frac{-1}{N}\sum^{N}_{n=1}\log \mathcal{A}(\bm{x}_n,\bm{y}_n)
\end{equation}
and
\begin{equation}
    \label{eq:la0}
    \mathcal{L}_{A}^{0}(\bm{X},\bm{Y}) = \frac{-1}{N}\sum^{N}_{n=1} \log(1-\mathcal{A}(\bm{x}_n,\argmax_{\bm{y}}\mathcal{M}(\bm{x}_n))) 
\end{equation}
where $\mathcal{A}(\cdot)$ is the output of the \emph{A-net} and $\mathcal{M}(\cdot)$ is the output of the \emph{M-net}.

Hence $\mathcal{L}_A(\bm{X},\bm{Y})$ simplifies to
\begin{equation}
    \label{eq:lossA}
    \mathcal{L}_A({\bm{X},\bm{Y}}) = 
    \frac{-1}{2N} \sum^{N}_{n=1} \log \mathcal{A}(\bm{x}_n,\bm{y}_n)
    +\; \log(1-f_A(\bm{x}_n,\argmax_{\bm{y}}\mathcal{M}(\bm{x}_n)))
\end{equation}

On the other hand, the
main network \emph{M-net} performs the actual set of \emph{tasks} we aim at. In the 
\emph{M-net} the cost function is composed by the contribution of two cost functions, 
whose balance is controlled by the hyperparameter $\lambda$
\begin{equation}
    \mathcal{L}_M(\bm{X},\bm{Y}) = \frac{-1}{N} \sum^{N}_{n=1} 
    \mathcal{L}(\bm{x}_n,\bm{y}_n) + \lambda 
    \mathcal{L}_{A}^{0}(\bm{x}_n,\bm{y}_n) 
\end{equation}
where $\mathcal{L}^{0}_A(\cdot)$ (Eq.~\ref{eq:la0}) drives the network to fool the network \emph{A}, and 
$\mathcal{L}(\bm{x}_i,\bm{y}_i)$ is the cross-entropy loss, which drives the network to 
learn the probability distribution of the training data:
\begin{equation}
    \label{eq:lossM}
    \mathcal{L}(\bm{x},\bm{y}) = \frac{-1}{Twh} 
    \sum^{T}_{t=1}\sum^{w}_{i=1}\sum^{h}_{j=1}
    \sum^{K^{t}}_{c=1}\bm{y}^{t}_{i,j,c}
    \log(\mathcal{M}^{t}_{i,j,c}(\bm{x}))
\end{equation}
where $T$ is the number of tasks to be performed, $K^{t}$ is the number of classes 
of the task $t, 1 \le t \le T$, the binary target variable $\bm{y}^{t}_{i,j,c} \in \{0, 1\}$ has a
1-of-$K^{t}$ coding scheme indicating the correct class, and $\mathcal{M}^{t}_{i,j,c}(x)$ is 
the output of the \emph{M-net} network for the \emph{Task-}$t$ and class $c$, 
interpreted as $\mathcal{M}^{t}_{i,j,c}(\bm{x})=\mathrm{p}(\bm{y}^{t}_{i,j,c}~=~1|\bm{x}_{i,j})$ (i.e. the posterior
probability of the pixel $i,j$ of the input image belongs to the class $c$ for task $t$).

Both ANNs are optimized in parallel, following the standard approach from \cite{goodfellow2014generative}:
we alternate between one gradient descent step on \emph{M-net}, and one step on \emph{A-net},
and so on. 

The training set-up is depicted in Fig.~\ref{fig:DAN}, where the ``Mux'' block is a 
standard multiplexer between the ``real'' label ($\bm{y}_n$) and the ``fake'' one
from the \emph{M-net}.

\subsection{Stage 2: Zones Segmentation and baseline detection}
On this stage information extracted from the images in previous stage is shaped
in a useful way, into a set piece-wise linear curves and polygons.
\subsubsection{Contour Extraction}
Let a test instance $\bm{x}$ and its pixel level classification $\bm{y}^*$ 
obtained in the previous stage be given. First, the contour extraction algorithm 
presented by Suzuky et al. \cite{suzuki1985topological} is used for each
zone over $\bm{y}^2$ to determine the vertices of its contour. This algorithm provides a set of 
contours $\Phi_k=\{\phi_1, \phi_2,...,\phi_R\},k \in \mathcal{Y}^2, R \ge 0$. 
Then, for each contour in $\Phi_k$ the same extraction algorithm is used over 
$\bm{y}^1$ to find the contours where baselines are expected to be, but restricted to
the area defined by the contour 
$\phi_r, 1 \ge r \ge R$. In this step, a new set of contours 
$L_{k,r}={l_1,l_2,\dots,l_S}; S \ge 0$ are found.

Finally, each contour $l_s,\,0\!\ge\!s\!\ge\!S$ is supposed to contain a single line
of text, whereby a simple baseline detection algorithm is applied to the section
of the input image within the contour $l_s$ (see Sec.~\ref{sec:baseline_alg}).

Notice that \emph{Task-1} and \emph{Task-2} can be treated independently using the 
same formulation above by simply ignoring the network output associated to the 
task we are not interested in. Then, in Stage 2 we set the regions of interest to be only 
one with size equal to the input image, as defined in Eq.~(\ref{eq:noTB}), to 
perform \emph{Task-1} only. 

\begin{equation}
	\label{eq:noTB}
	\Phi_k=\Phi_1=\{\phi_1\}; \phi_1=[(0,0), (w,0), (w,h), (0,h)]
\end{equation}

Similarly to perform \emph{Task-2} alone, we just return $\Phi_k, \forall k \in \mathcal{Y}^2$, without further 
search inside those contours.

\subsubsection{Baseline detection algorithm}\label{sec:baseline_alg}
Once a text line contour $l_s$ is detected, a very simple algorithm can be used 
to detect the baseline associated to that specific text line (under the 
assumption that there is only one text line per contour). 
Each baseline is first straightfordwardly represented as a digital curve.

The pseudo code of the algorithm is presented on Alg.~\ref{alg:base_det}.
First the input image $\bm{x}$ is cropped with the polygon defined in $l_s$, and 
it is binarized using Otsui's algorithm. Then, we define the lowest black pixel 
of each column in the binarized image as a point of the digital curve we are 
searching for. Finally, as a result of Alg.~\ref{alg:base_det} line 7, the number of points of the 
digital curve $\rho$ is equal to the number of columns of the cropped 
image $Y$. In order to reduce the number of points of $\rho$ and remove some outliers, the 
algorithm presented by Perez~et~al.~\cite{perez1994} to find an optimal 
piece-wise linear curve with only $m$ vertices is used.

\begin{algorithm}[!t]
	\KwData{an image $\bm{x}$, a contour $l_s$, number of vertex of output 
    piece-wise linear curve $m$}
	\KwResult{piece-wise linear curve $v$}
 	$I=crop(\bm{x},l_s)$ \\
    $Y=Otsu(I)$ \\
    $\rho = []$\\
    \For{i in rows($Y$)}{
    	\For{j in columns($Y$)}{
        	\If{$Y_{i,j} == 1$}{
            	$\rho[j] = (j,i)$ 
            }
        }
    }
    $v$ = reducePoly($\rho$,$m$) 
\caption{Baseline detection algorithm}
\label{alg:base_det}
\end{algorithm}


\section{Experimental Set-up}\label{sec:exp}
To assess the performance of the proposed method we test it on three publicly
available datasets: cBAD\footnote{\url{https://zenodo.org/record/257972}}, 
Bozen\footnote{\url{https://zenodo.org/record/1297399}}, and a new dataset called
\textit{Oficio de Hipotecas de Girona} (OHG)\footnote{\url{https://zenodo.org/record/1322666}}.

OHG is a new dataset introduced in~\cite{Quiros2018_b} for HTR and here for
DLA. The ground-truth is annotated both with baselines and
zones (segmentation and several labels) (details on~\ref{sec:girona}), which 
will allows us to carry out more detailed 
experiments with complex layout images.

All experiments are conducted using the same hardware, a single NVIDIA TitanX GPU 
installed along with an Intel Core i5-2500K@3.30GHz CPU with 16GB RAM. The 
source code, along with the configuration files to replicate these experiments, 
are available at \url{https://github.com/lquirosd/P2PaLA}.

\subsection{Ground-truth}
Ground-truth is recorded in PAGE-XML format because it allows us to manually annotate 
and review the elements (baselines and zones) easily, as they can be
defined by piece-wise linear curve or polygons of just few vertices.

The ground-truth is then processed to encode the data into the 1-of-$K^t$ coding scheme 
in order to train the ANNs~( Eq.~\ref{eq:lossM}).
An example of this ground-truth is 
shown in Fig.~\ref{fig:label}, where each color represents a different value 
in the encoding.

\subsection{Artificial Neural Network Architecture}
As mentioned in Sec.~\ref{sec:stage1}, the proposed ANN architecture is very
similar to the one presented by \cite{pix2pix2016}, but it was modified to 
perform a discriminative rather than a generative processing. The main hyper-parameters 
of each part of the ANN are reported below, following the 
convention presented in \cite{pix2pix2016}, where \texttt{Ck} denotes a 
Convolution-BatchNorm-LeakyReLU layer with \texttt{k} filters, and \texttt{CDk}
denotes a Convolution-BatchNorm-Dropout-ReLU layer with a dropout rate of 0.5.

\subsubsection{\emph{A-net} Network Architecture}
This network is a simple single output Convolutional Neural Network, trained as
explained in Sec.~\ref{sec:objective}.
Its main parameters are:

\begin{itemize}
    \item number of input channels: defined by the number of 
    channels of the input image (3 for RGB images) plus one more for each task 
    involved. In the case of two tasks and RGB images number of input channels 
    is 5.
    \item Architecture: \texttt{C64-C128-C256-C512-C512-C512-Sigmoid}.
    \item Convolution filters: $4\times 4$, stride 2.
\end{itemize}

\subsubsection{\emph{M-net} Network Architecture}
This network is structured as an encoder-decoder architecture called
U-Net~\cite{DBLP:RonnebergerFB15}. 
U-Net differs from a common  encoder-decoder due its skip connections between 
each layer $i$ in the encoder and layer $n-i$ in the decoder, where $n$ is the 
total number of layers. Main parameters are:
\begin{itemize}
\item number of input channels: defined by the number of channels of the 
input image (3 for RGB images).
\item Architecture:
\begin{itemize}
\item encoder: \texttt{C64-C128-C256-C512-C512-C512-C512-C512}
\item decoder: \texttt{CD512-CD1024-CD1024-C1024-C1024-C512-C256-C128-SoftMax}, where LeakyReLU layers are changed to ReLU.
\item Convolution filters: $4\times 4$, stride 2.
\end{itemize}
\end{itemize}

\subsubsection{Training and Inference}
To optimize the networks we follow \cite{pix2pix2016}, using minibatch SGD and Adam 
solver \cite{kingma2014adam}, with a learning rate of 0.0001, and momentum parameters 
$\beta_1=0.5$ and $\beta_2=0.999$. Also, we use weighted loss from 
\cite{paszke2016enet}, to overcome the imbalance problem in \emph{Task-2}. The weight is computed as 
$w_k = \frac{1}{\log(c+p_k)}, k \in \mathcal{Y}^t, c \geq 0$, where $p_k$ is the
prior-probability of the $k$-th value associated with the task.

Affine transformations (translation, rotation, shear, scale) and Elastic 
Deformations~\cite{Simard03} are applied to the input images as a data augmentation technique,
where its parameters are selected randomly from a restricted set of allowed values, 
and applied on each epoch and image with a probability of 0.5.

In our experiments, we use the maximum batch size allowed by the hardware 
we have available: 8 images of size $1024\times 768$ on a single Titan X GPU.

\subsection{Evaluation Measures}

\subsubsection{Baseline Detection}
We report precision (P), recall (R) and its harmonic mean (F1) measures as defined 
specifically for this kind of problem 
in \cite{Gruning2017}. Tolerance parameters are set to default values in all 
experiments (see \cite{Gruning2017} for details about measure definition, tolerance
values and implementation details).

\subsubsection{Zone Segmentation}
We report metrics from semantic segmentation and scene parsing evaluations as
presented in \cite{long2015fully}:
\begin{itemize}
\item Pixel accuracy (pixel acc.): ${\sum_i \eta_{ii}}/{\sum_i \tau_i}$
\item Mean accuracy (mean acc.): $1/K^{t=2}\sum_i \eta_{ii}/\tau_i$
\item Mean Jaccard Index (mean IU): $(1/K^{t=2})\sum_i \eta_{ii}/(\tau_i+\sum_j \eta_{ji}-\eta_{ii})$
\item Frequency weighted Jaccard Index (f.w. IU): $(\sum_\kappa \tau_\kappa)^{-1} \sum_i \tau_i \eta_{ii}/(\tau_i+\sum_j \eta_{ji}-\eta_{ii})$
\end{itemize}
where $\eta_{ij}$ is the number of pixels of class $i$ predicted to belong to class
$j$, $K^{t=2}$ is the number of different classes for the task $t=2$, $\tau_i$ the number of pixels
of class $i$, and $\kappa \in \mathcal{Y}^{t=2}$.

\subsection{Data Sets}

\subsubsection{\textit{Oficio de Hipotecas de Girona}}\label{sec:girona}

\begin{figure}[htbp]
    \centering
    \includegraphics[width=0.8\linewidth]{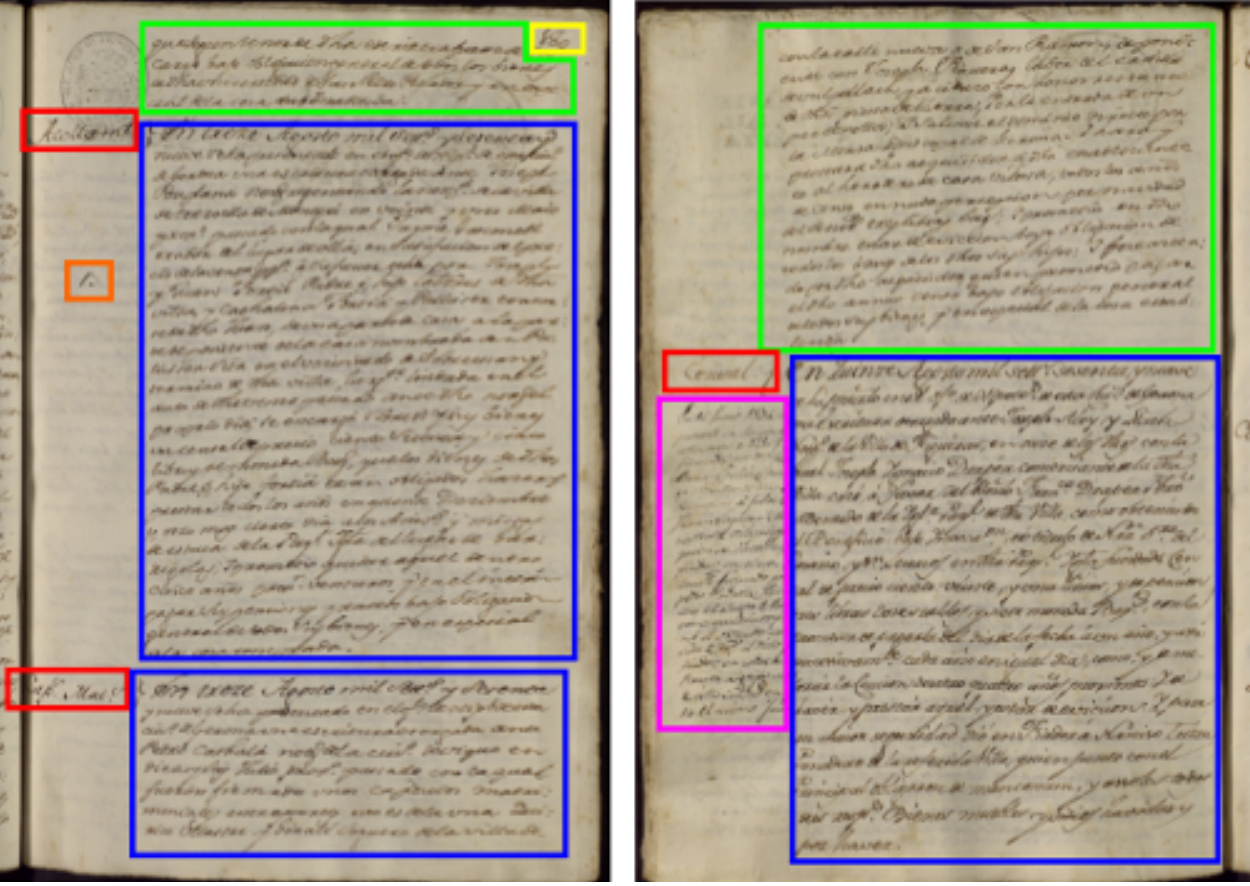}
    \caption{Examples of pages with different layouts, belonging to the
    \textit{Oficio de Hipotecas de Girona} dataset.
    Yellow: \texttt{\$pag}, red: \texttt{\$tip}, green: \texttt{\$pac}, 
    blue: \texttt{\$par}, fuchsia: \texttt{\$not}, 
    orange:\texttt{\$nop} (better seeing in color).}
    \label{fig:girona_page}
\end{figure}

The manuscript \textit{Oficio de Hipotecas de Girona} (OHG) 
is provided by the Centre de Recerca d'Hist\`oria Rural from 
the Universitat de Girona (CRHR)\footnote{\url{http://www2.udg.edu/tabid/11296/Default.aspx}}.
This collection is composed of hundreds of 
thousands of  notarial deeds from the XVIII-XIX century (1768-1862) \cite{Quiros2018}. Sales, 
redemption of censuses, inheritance and matrimonial chapters are among the 
most common documentary typologies in the collection. This collection is divided
in batches of 50 pages each, digitized at 300ppi in 24 bit RGB color, available 
as TIF images along with their respective ground-truth layout in PAGE XML format,
compiled by the HTR group of the PRHLT\footnote{\url{https://prhlt.upv.es}} center
and CRHR. OHG pages exibit a relatively complex layout, composed of six relevant 
zone types; namely: 
\texttt{\$pag, \$tip, \$par, \$pac, \$not, \$nop},
as described in Table~\ref{tab:layout}. An
example is depicted in Fig.~\ref{fig:girona_page}.

\begin{table}[htbp]
    \centering
    \caption{Layout regions in the OHG dataset.}
    \label{tab:layout}
    \begin{tabular}{ll}
        \toprule
        ID & Description \\
        \midrule
        \texttt{\$pag} & page number.\\
        \texttt{\$tip} & notarial typology.\\
        \texttt{\$par} & a paragraph of text that begins next to a notarial typology.\\
        \texttt{\$pac} & a paragraph that begins on a previous page.\\
        \texttt{\$not} & a marginal note.\\
        \texttt{\$nop} & a marginal note added a posteriori to the document.\\
        \bottomrule
    \end{tabular}
\end{table}

In this work we use a portion of 350 pages from the collection, from batch 
\emph{b004} to batch \emph{b010}, divided ramdomly into training and test sets,
300 pages and 50 pages respectively. Main characteristics of this dataset are 
summarized on Table \ref{tab:corpus}.

\begin{table}[ht]
\centering
\caption{Main characteristics of the OHG dataset.}
\label{tab:corpus}
\begin{tabular}{@{}cccccccc@{}} \toprule
\multirow{2}{*}{Batch} & \multirow{2}{*}{\#Lines} & \multicolumn{6}{c}{\#Zones}\\ 
\cmidrule(r){3-8}
    & & \texttt{\$par} & \texttt{\$pac} & \texttt{\$tip} & \texttt{\$pag} & \texttt{\$nop} & \texttt{\$not}\\ 
\midrule
b004 & 1960 & 72 & 35 & 67 & 24 & 28 & 6 \\
b005 & 1985 & 73 & 41 & 71 & 25 & 31 & 2 \\
b006 & 1978 & 68 & 42 & 68 & 25 & 24 & 4 \\
b007 & 1762 & 60 & 33 & 62 & 19 & 26 & 1 \\
b008 & 1963 & 69 & 39 & 69 & 24 & 30 & 3 \\
b009 & 1976 & 75 & 40 & 75 & 25 & 34 & 2 \\
b010 & 2023 & 71 & 38 & 71 & 25 & 43 & 3 \\
\midrule
Total & 13647 & 488 & 268 & 483 & 167 & 216 & 21 \\

\bottomrule
\end{tabular}
\end{table}

\subsubsection{cBAD dataset}

This dataset was presented in~\cite{diem2017} for the ICDAR 2017 Competition on 
Baseline Detection in Archival Documents (cBAD).
It is composed of 
2035 annotated document page images that are collected from 9 different archives.
Two competition tracks an their corresponding partitions are defined on this 
corpus to test different characteristics of 
the submitted  methods. Track A [Simple Documents] is published with annotated 
text regions and therefore aims to evaluate the quality of text line segmentation 
(216 pages for training and 539 for test). 
The more challenging Track B [Complex Documents] provides only the page area 
(270 pages for training and 1010 for test). 
Hence, baseline detection algorithms need to correctly locate text lines in the
presence of marginalia, tables, and noise. The dataset comprises images with 
additional PAGE XMLs, which contain text regions and baseline annotations.

\subsubsection{Bozen dataset}

This dataset consists of a subset of documents from the Ratsprotokolle collection
composed of minutes of the council meetings held from 1470 to 1805 (about 30.000
pages)\cite{sanchez2016}. The dataset text is written in Early Modern German by an
unknown number of writers.
The public dataset is composed of 400 pages (350 for training and 50 for 
validation); most of the pages consist of a two or three zones with many difficulties 
for line detection and extraction.

\section{Results}\label{sec:results}

\subsection{\textit{Oficio de Hipotecas de Girona}}

The dataset is divided randomly into a training and a test set, 300 pages and 50 pages
respectively. Experiments are conducted on incremental training subsets from 16 to 300 
training images, for \emph{Task-1} and \emph{Task-2}.

Two experiments are performed using this dataset. First, the system is
configured to perform only \emph{Task-1} giving as a result only the baselines 
detected in the input images. Second, the system is
configured to perform both tasks in a integrated way, giving as a result both the baselines and
the layout zones (both segmentation and labels).

\begin{figure}
    \centering
    \includegraphics[width=\linewidth]{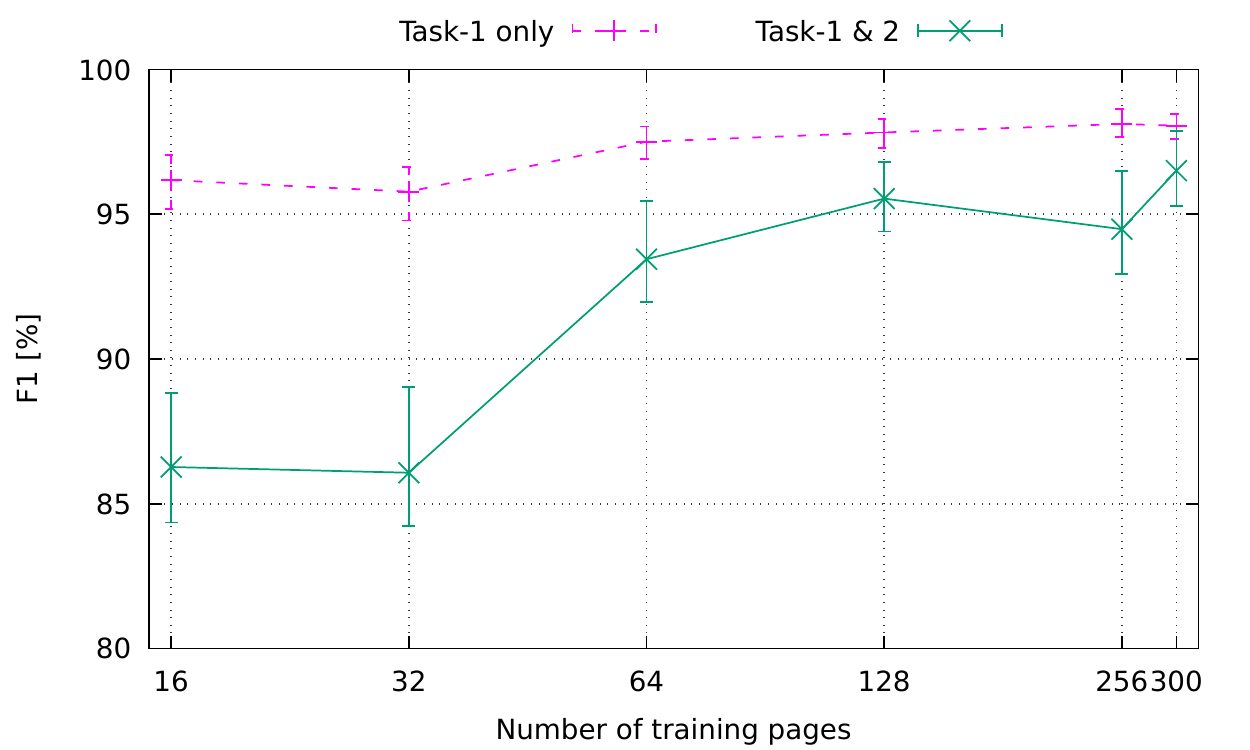}
    \caption{F1 measure for OHG using different number of training pages.
        \emph{Task-1} defined as baseline detection only, and \emph{Task-1} \& \emph{2} as 
    baseline detection plus zone segmentation and labeling.}
    \label{fig:girona_results_bl}
\end{figure}

\begin{figure}
    \centering
    \includegraphics[width=\linewidth]{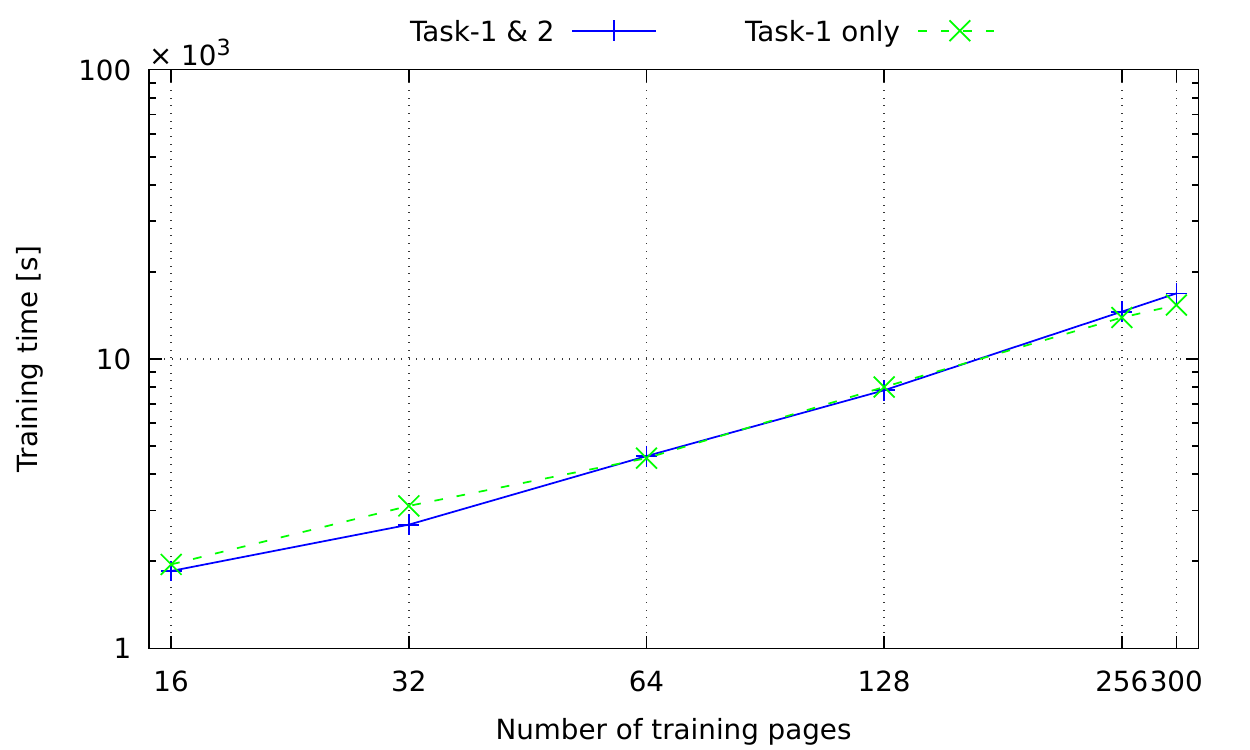}
    \caption{Training time required on OHG experiments. The trend is linear with
    respect to the number of pages. Better seeing in color.}
    \label{fig:girona_time}
\end{figure}

Baseline detection precision and recall results for both experiments are reported in Table~\ref{tab:results_OHG}
and F1-measure is depicted in Fig.~\ref{fig:girona_results_bl}.
Even though there are statistically significant differences between the results of performing 
\emph{Task-1} alone and performing both tasks, the slight degradation when both 
tasks are solved simultaneously is admissible 
because of the benefit of having not only the baselines detected, but also the 
zones segmented and labeled. Moreover,
when we have enough training images the F1 difference becomes small. Also, there 
is no appreciable impact in the
training time required by the system when we use two tasks or only one (see 
Fig.~\ref{fig:girona_time}).

\begin{table}
    \centering
    \caption{Precision (P) and Recall (R) results for \emph{Task-1} of OHG, when system is trained for 
    \emph{Task-1} only and when is trained for both tasks, using different number of training pages.
    Nonparametric Bootstrapping confidence intervals at 95\%, 10000 repetitions are provided.}
    \label{tab:results_OHG}
    \begin{tabular}{lcc|cc}
        \toprule
        \multirow{2}{*}{\shortstack[l]{\# of \\pages}} & \multicolumn{2}{c}{Both tasks} & \multicolumn{2}{c}{\emph{Task-1} only}\\
                                  & P & R                             & P & R \\
        \midrule
        16 & 81.5~[78.6,84.1] & 92.6~[90.9,94.1] & 93.3~[95.1,97.3] & 96.0~[95.2,96.7] \\
        32 & 79.6~[76.1,83.1] & 95.1~[94.3,95.8] & 96.2~[95.3,97.1] & 95.3~[94.2,96.2] \\
        64 & 91.8~[89.1,94.3] & 95.9~[95.1,96.6] & 97.5~[96.9,98.1] & 97.5~[96.9,97.9] \\ 
        128& 94.8~[93.1,96.4] & 96.5~[95.7,97.1] & 98.0~[97.5,98.4] & 97.6~[97.1,98.1] \\
        256& 93.3~[90.2,95.9] & 96.4~[95.7,97.1] & 98.2~[97.8,98.6] & 98.0~[97.5,98.6] \\       
        300& 96.2~[94.1,97.9] & 97.1~[96.4,97.7] & 98.4~[98.1,98.7] & 97.7~[97.2,98.1] \\
        \bottomrule
    \end{tabular}
\end{table}

\begin{figure}
    \centering
    \includegraphics[width=0.8\linewidth]{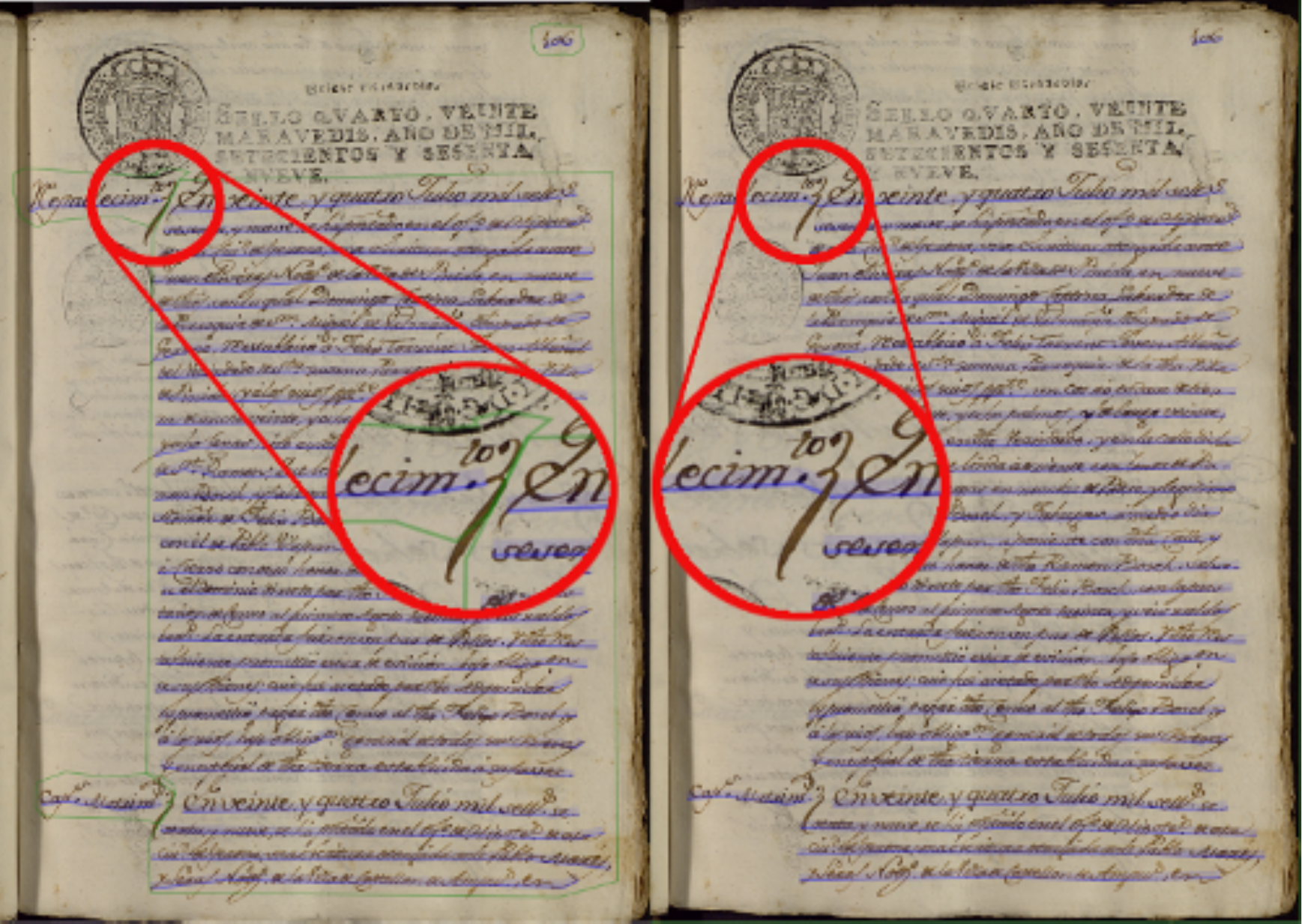}
    \caption{Example of OHG results. In the left zone segmentation prevents the 
    baselines to be merged (\emph{Task-1} and \emph{Task-2} are performed in an 
    integrated manner). In the right the baselines are merged (\emph{Task-1} is 
    performed alone). Better seeing in color.}
    \label{fig:girona_example}
\end{figure}

\begin{figure}
    \centering
    \includegraphics[width=0.8\linewidth]{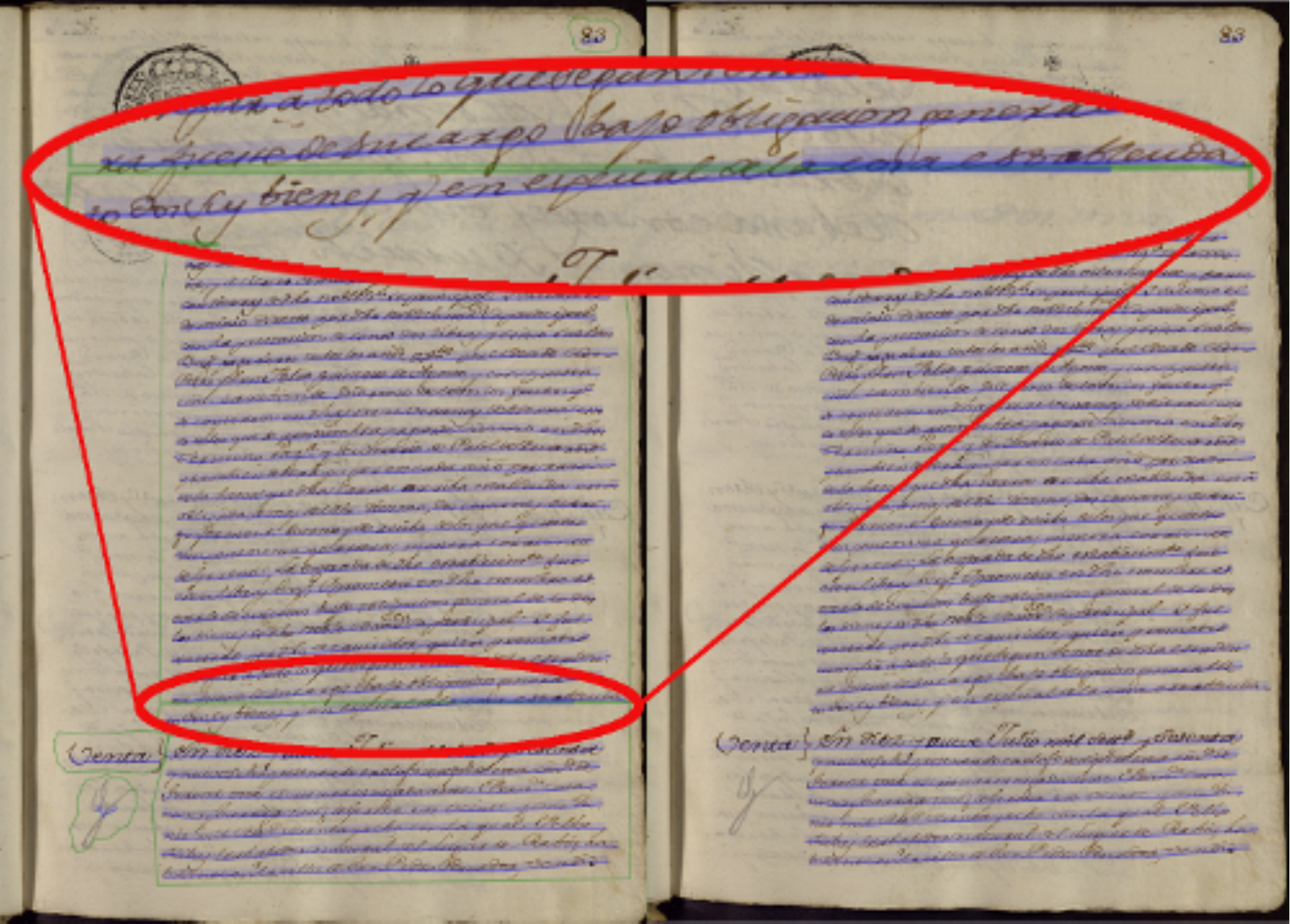}
    \caption{Example of OHG results, where zone segmentation forces 
    baseline to split. Left side is an example where 
    \emph{Task-1} and \emph{Task-2} are performed, right side is an example when only
    \emph{Task-1} is performed. As remarked, a baseline is divided into two at 
    the point where intersects the zone boundary. Better seeing in color.}
    \label{fig:girona_example_w}
\end{figure}

The recall measure obtained in both experiments is very stable across the number of 
training images, while precision is closely related to the quality of the 
zones segmented in the \emph{Task-2}, see Fig.~\ref{fig:girona_example} and 
\ref{fig:girona_example_w}.

\begin{figure}
    \centering
    \includegraphics[width=\linewidth]{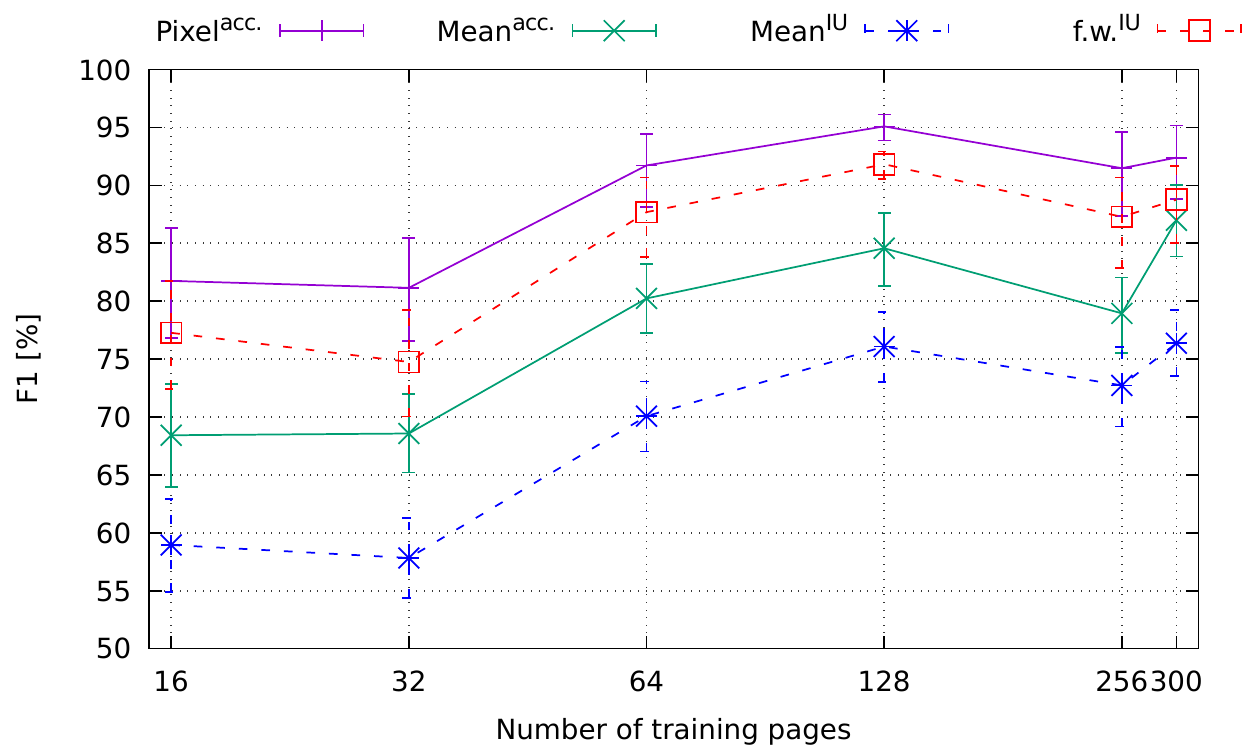}
    \caption{Results for \emph{Task-2} of OHG.
    Nonparametric Bootstrapping confidence intervals at 95\%, 10000 repetitions.
    Better seeing in color.}
    \label{fig:girona_results_zones}
\end{figure}

Zone segmentation results are reported on Fig.~\ref{fig:girona_results_zones}.
As expected, an improvement with increasing number of training images is observed until 128
images. There the results keep varying but without significant statistical difference. 

\subsection{cBAD}
For this work, only Track B documents are used to train the system. The ground-truth of the test 
set is not available to the authors, whereby metrics are computed through the
competition website\footnote{https://scriptnet.iit.demokritos.gr/competitions/5/}.

The system was trained through 200 epochs to perform \emph{Task-1} only, because
no ground-truth is available for Text Zones in the dataset. Training time was
around 3.75 hours using 270 training images on a mini-batch of 8.

Results are reported in Table~\ref{tab:cbad_results}, along with state-of-the-art 
results presented in the competition and two others recently published (dhSegment,
ARU-Net). The proposed approach achieved very 
competitive results on such a heterogeneous dataset, without significant statistical
difference with respect to the winner method of the competition (DMRZ). But below 
ARU-Net latest result, which we believe is mainly due to the simple baseline detection 
algorithm we used in \textit{Stage 2}.

\begin{table}[htbp]
	\centering
    \caption{Precision (P) and Recall (R) results for the cBAD test set (\emph{Task-1} only).
    Nonparametric Bootstrapping confidence intervals at 95\%, 10000 repetitions.}
    \label{tab:cbad_results}
	\begin{tabular}{lccc}
	\toprule
    Method & P & R & F1 \\
    \midrule
    IRISA    & 69.2 & 77.2 & 73.0 \\ 
    UPVLC    & 83.3 & 60.6 & 70.2 \\
    BYU      & 77.3 & 82.0 & 79.9 \\
    proposed & 84.8 [83.9, 85.7] & 85.4 [84.4, 86.4] & 85.1 \\ 
    DMRZ     & \textbf{85.4} & \textbf{86.3} & \textbf{85.9} \\ 
    dhSegment~\cite{Ares2018}  & 82.6 & 92.4 & 87.2 \\
    ARU-Net~\cite{Gruning2018}  & \textbf{92.6} & \textbf{91.8} & \textbf{92.2} \\
    \bottomrule
	\end{tabular}
\end{table}

\begin{figure}
    \centering
    \includegraphics[width=0.8\linewidth]{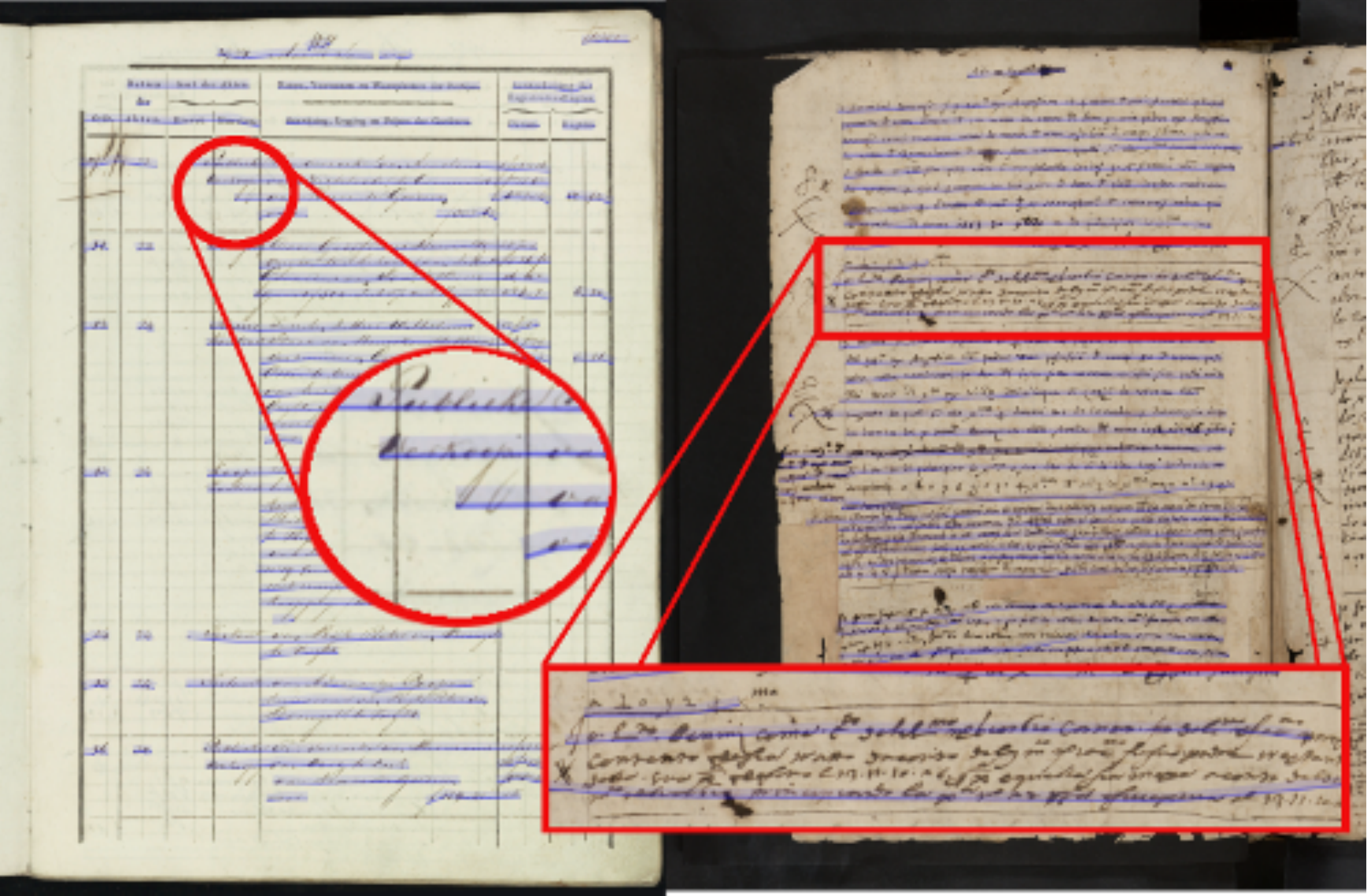}
    \caption{Example of errors in cBAD results produced by the proposed approach.
    Merged baselines are shown in the left, 
    where adjacent lines in a table are detected as a single one.
    Missing baselines are showed in the right. Better seeing in color.}
    \label{fig:cbad_results}
\end{figure}

Main errors are related to merged baselines or missing lines in very crowded areas.
An example of those errors is shown in Fig.~\ref{fig:cbad_results}.

\subsection{Bozen}
Experiments on this work are conducted using the training/validation splits defined
by the authors of the dataset, as training and test respectively.

The system was trained through 200 epochs to perform tree different experiments:
(I) only \emph{Task-1}, (II) integrated \emph{Task-1} and \emph{Task-2} and
(III) only 
\emph{Task-2}. Training time  for each experiment was 
around 4.75 hours using 350 training images and a mini-batch of 8.  

A F1 measure of 97.4\% has been achieved
on experiment (I), while results achieved on experiment (II) have no significant 
statistical difference (as shown in Table~\ref{tab:bozen_results}) but with the 
benefit of obtaining the zones.

Results of experiment (I) can be compared with \cite{Gruning2018} where a 97.1\% 
F1 measure is reported, which have no significant statistical difference with the
results reported here. 

\begin{table}[htbp]
    \centering
    \caption{ 
    Precision (P), Recall (R), F1, Pixel accuracy (Pixel\textsuperscript{acc.}),
    Mean Pixel accuracy (Mean\textsuperscript{acc.}), Mean Jaccard Index (Mean\textsuperscript{IU}) and
    Frequency weighted Jacard Index (f.w.\textsuperscript{IU}) results for the Bozen test set.
    Nonparametric Bootstrapping confidence intervals at 95\%, 10000 repetitions.}
    \label{tab:bozen_results}
    \begin{tabular}{p{15mm}ccc}
        \toprule
         Metric & \emph{Task-1} only & \emph{Task-1 and 2} & \emph{Task-2} only \\
         \midrule
         \multicolumn{4}{c}{Baseline Detection} \\
         \midrule
         P~[\%]  & 95.8~[92.7, 97.8] & 94.5~[92.9, 95.9] & -- \\
         R~[\%]  & 99.1~[98.6, 99.4] & 98.9~[98.5, 99.3] & -- \\
         F1~[\%] & 97.4              & 96.6 & -- \\

         \midrule
         \multicolumn{4}{c}{Zone Segmentation} \\
         \midrule
         Pixel\textsuperscript{acc.}~[\%]  & -- & 95.5~[94.8, 96.1] & 95.3~[94.6, 96.0] \\
         Mean\textsuperscript{acc.}~[\%]   & -- & 91.4~[90.1, 92.7] & 93.3~[92.1, 94.5] \\
         Mean\textsuperscript{IU}~[\%]     & -- & 84.5~[83.1, 85.8] & 82.7~[81.3, 84.1] \\
         f.w.\textsuperscript{IU}~[\%]     & -- & 91.6~[90.5, 92.6] & 91.3~[90.2, 92.4] \\
        \bottomrule
    \end{tabular}
\end{table}

\begin{figure}
    \centering
    \includegraphics[width=0.8\linewidth]{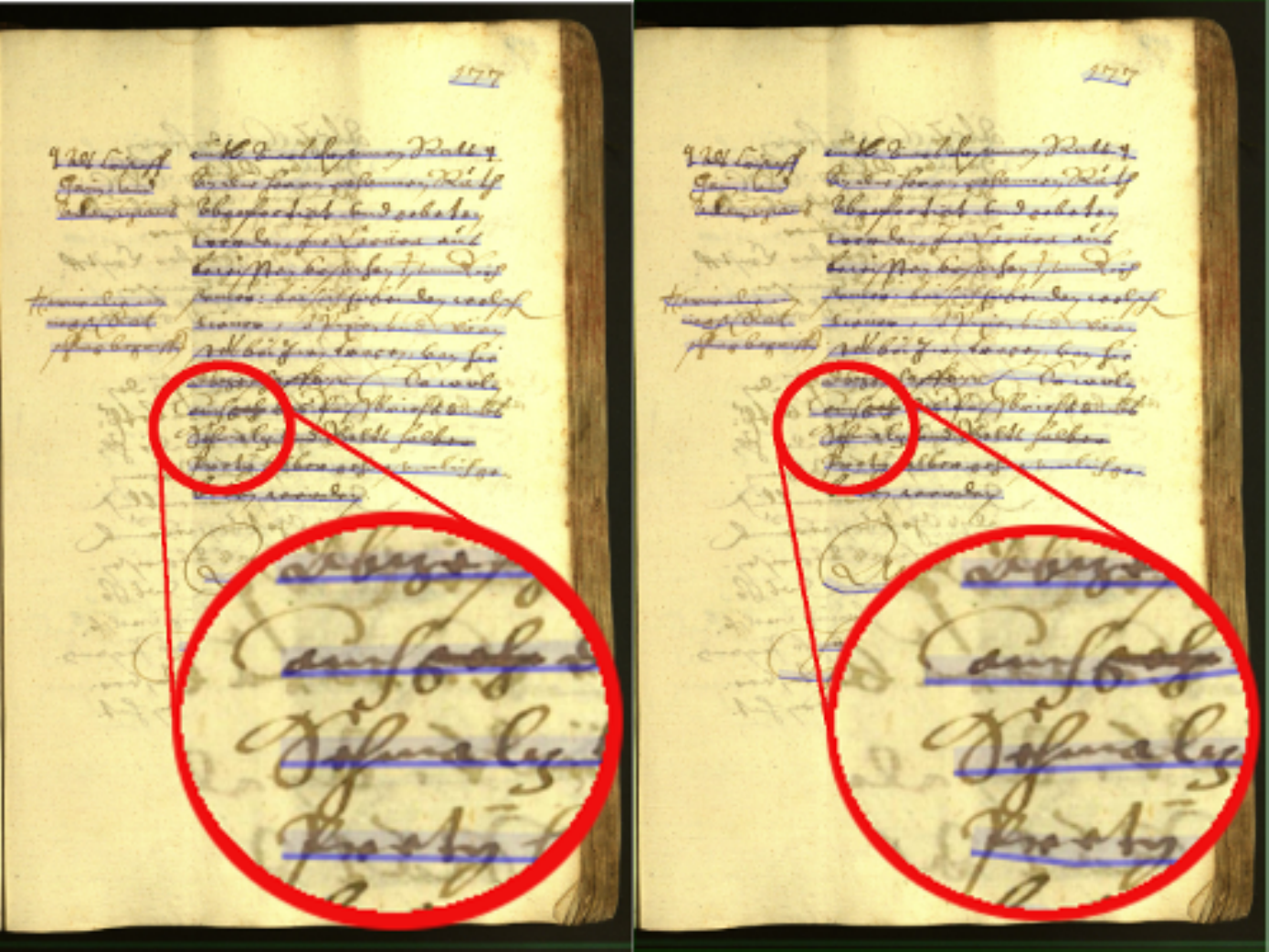}
    \caption{Example of Bozen result. Left image is the ground-truth and at the 
    right is the output of the system. Most of the differences are found at 
    the beginning or the end of the baseline.
    Better seeing in color.}
    \label{fig:bozen_results}
\end{figure}

An example of the errors obtained in this experiments  
is shown in Fig.~\ref{fig:bozen_results}, where those differences do not generally affect
the results of subsequent HTP systems.

Zone segmentation and labeling results of experiments (II) and (III) indicate that
there is no significant loss in the quality of the results obtained when the system is trained
to perform only one of the tasks or both integrated. On the other hand, the average
computation time per page at test is reduced by 68\% (1.13~s and 0.36~s 
respectively) as expected.

%

\section{Conclusions}\label{sec:conclusions}
In this paper we present a new multi-task method for handwritten document layout analysis,
which is able to perform zone segmentation and labeling, along with baseline detection, in a 
integrated way, using a single model. The
method is based on discriminative ANN and a simple contour and baseline detection algorithms.

We conducted experiments in three different datasets, with promising results on all
of them without model reconfiguration or hyper-parameter tuning. 

The integrated model
tasks) the ANN parameters across tasks without significant degradation in the quality of
the results.

Baseline detection
results in OHG and Bozen are good enough for most HTR and KWS applications, while 
cBAD results may not be enough for HTR applications if high quality transcripts
are expected. In this sense, we will study the introduction of restrictions and 
prior probabilities in the optimization problem to prevent unfeasible hypothesis 
and reduce the searching space.
Also, we will explore the application of the Interactive Pattern
Recognition framework established in~\cite{Toselli2011} for layout 
analysis~\cite{Quiros2017} to help users to easily review
the document layout before feeding the results to the HTP system.


\section*{Acknowledgments}

The author would like to acknowledge Alejandro H. Toselli, Carlos-D. Martınez-Hinarejos and
Enrique Vidal for their reviews and advice. 
NVIDIA Corporation kindly donated the Titan X GPU used for this
research. Finally, this work was partially supported by the
Universitat Polit\`ecnica de Val\`ecia under grant FPI-II/899, a 2017-2018 
Digital Humanities research grant of the BBVA Foundacion for the project
"Carabela", and through the
EU project READ (Horizon-2020 program, grant Ref. 674943).

\bibliographystyle{elsarticle-num}
\bibliography{P2PaLA}

\begin{thebibliography}{10}
\expandafter\ifx\csname url\endcsname\relax
  \def\url#1{\texttt{#1}}\fi
\expandafter\ifx\csname urlprefix\endcsname\relax\def\urlprefix{URL }\fi
\expandafter\ifx\csname href\endcsname\relax
  \def\href#1#2{#2} \def\path#1{#1}\fi

\bibitem{romero2011}
V.~Romero, N.~Serrano, A.~H. Toselli, J.~A. S{\'a}nchez, E.~Vidal, Handwritten
  text recognition for historical documents, in: Proc. of the Workshop on
  Language Technologies for Digital Humanities and Cultural Heritage, Hissar,
  Bulgaria, 2011, pp. 90--96.

\bibitem{Bluche2017}
T.~Bluche, S.~Hamel, C.~Kermorvant, J.~Puigcerver, D.~Stutzmann, A.~H. Toselli,
  E.~Vidal, Preparatory kws experiments for large-scale indexing of a vast
  medieval manuscript collection in the himanis project, in: 2017 14th IAPR
  International Conference on Document Analysis and Recognition (ICDAR),
  Vol.~01, 2017, pp. 311--316.
\newblock \href {http://dx.doi.org/10.1109/ICDAR.2017.59}
  {\path{doi:10.1109/ICDAR.2017.59}}.

\bibitem{fornes2017}
A.~Forn{\'e}s, V.~Romero, A.~Bar{\'o}, J.~I. Toledo, J.~A. S{\'a}nchez,
  E.~Vidal, J.~Llad{\'o}s, Icdar2017 competition on information extraction in
  historical handwritten records, in: Document Analysis and Recognition
  (ICDAR), 2017 14th IAPR International Conference on, Vol.~1, IEEE, 2017, pp.
  1389--1394.

\bibitem{cattoni1998}
R.~Cattoni, T.~Coianiz, S.~Messelodi, C.~M. Modena, Geometric layout analysis
  techniques for document image understanding: a review, Tech. rep., ITC-irst
  (1998).

\bibitem{Romero2015}
V.~Romero, J.-A. S{\'a}nchez, V.~Bosch, K.~Depuydt, J.~Does, Influence of text
  line segmentation in handwritten text recognition, in: 13th International
  Conference on Document Analysis and Recognition (ICDAR), 2015.

\bibitem{Nagy2000}
G.~Nagy, Twenty years of document image analysis in pami, IEEE Transactions on
  Pattern Analysis and Machine Intelligence 22~(1) (2000) 38--62.
\newblock \href {http://dx.doi.org/10.1109/34.824820}
  {\path{doi:10.1109/34.824820}}.

\bibitem{Mao2003}
S.~Mao, A.~Rosenfeld, T.~Kanungo, Document structure analysis algorithms: a
  literature survey, in: Document Recognition and Retrieval X, Vol. 5010,
  International Society for Optics and Photonics, 2003, pp. 197--208.

\bibitem{Namboodiri2007}
A.~M. Namboodiri, A.~K. Jain, Document structure and layout analysis, in:
  Digital Document Processing, Springer, 2007, pp. 29--48.

\bibitem{Eskenazi2017}
S.~Eskenazi, P.~Gomez-Kr{\"a}mer, J.-M. Ogier,
  \href{http://www.sciencedirect.com/science/article/pii/S0031320316303399}{A
  comprehensive survey of mostly textual document segmentation algorithms since
  2008}, Pattern Recognition 64 (2017) 1 -- 14.
\newblock \href
  {http://dx.doi.org/https://doi.org/10.1016/j.patcog.2016.10.023}
  {\path{doi:https://doi.org/10.1016/j.patcog.2016.10.023}}.
\newline\urlprefix\url{http://www.sciencedirect.com/science/article/pii/S0031320316303399}

\bibitem{Shi2009}
Z.~Shi, S.~Setlur, V.~Govindaraju, A steerable directional local profile
  technique for extraction of handwritten arabic text lines, in: 10th
  International Conference on Document Analysis and Recognition (ICDAR), 2009,
  pp. 176--180.
\newblock \href {http://dx.doi.org/10.1109/ICDAR.2009.79}
  {\path{doi:10.1109/ICDAR.2009.79}}.

\bibitem{Ryu2014}
J.~Ryu, H.~I. Koo, N.~I. Cho, Language-independent text-line extraction
  algorithm for handwritten documents, IEEE Signal Processing Letters 21~(9)
  (2014) 1115--1119.
\newblock \href {http://dx.doi.org/10.1109/LSP.2014.2325940}
  {\path{doi:10.1109/LSP.2014.2325940}}.

\bibitem{Ouwayed2012}
N.~Ouwayed, A.~Bela{\"i}d, \href{https://doi.org/10.1007/s10032-011-0172-6}{A
  general approach for multi-oriented text line extraction of handwritten
  documents}, International Journal on Document Analysis and Recognition
  (IJDAR) 15~(4) (2012) 297--314.
\newblock \href {http://dx.doi.org/10.1007/s10032-011-0172-6}
  {\path{doi:10.1007/s10032-011-0172-6}}.
\newline\urlprefix\url{https://doi.org/10.1007/s10032-011-0172-6}

\bibitem{Cohen2014}
R.~Cohen, I.~Dinstein, J.~El-Sana, K.~Kedem, Using scale-space anisotropic
  smoothing for text line extraction in historical documents, in: International
  Conference Image Analysis and Recognition, Springer, 2014, pp. 349--358.

\bibitem{Baechler2013}
M.~Baechler, M.~Liwicki, R.~Ingold, Text line extraction using dmlp classifiers
  for historical manuscripts, in: Document Analysis and Recognition (ICDAR),
  2013 12th International Conference on, IEEE, 2013, pp. 1029--1033.

\bibitem{Arvanitopoulos2014}
N.~Arvanitopoulos, S.~S{\"u}sstrunk, Seam carving for text line extraction on
  color and grayscale historical manuscripts, in: Frontiers in Handwriting
  Recognition (ICFHR), 2014 14th International Conference on, IEEE, 2014, pp.
  726--731.

\bibitem{Nicolaou2009}
A.~Nicolaou, B.~Gatos, Handwritten text line segmentation by shredding text
  into its lines, in: 10th International Conference on Document Analysis and
  Recognition (ICDAR), 2009, pp. 626--630.
\newblock \href {http://dx.doi.org/10.1109/ICDAR.2009.243}
  {\path{doi:10.1109/ICDAR.2009.243}}.

\bibitem{Bosch2012}
V.~Bosch~Campos, A.~H. Toselli, E.~Vidal, Natural language inspired approach
  for handwritten text line detection in legacy documents, in: Proceedings of
  the 6th Workshop on Language Technology for Cultural Heritage, Social
  Sciences, and Humanities, Association for Computational Linguistics, 2012,
  pp. 107--111.

\bibitem{Bosch2012_b}
V.~Bosch, A.~H. Toselli, E.~Vidal, Statistical text line analysis in
  handwritten documents, in: International Conference on Frontiers in
  Handwriting Recognition (ICFHR), 2012, pp. 201--206.
\newblock \href {http://dx.doi.org/10.1109/ICFHR.2012.274}
  {\path{doi:10.1109/ICFHR.2012.274}}.

\bibitem{Bosch2014}
V.~Bosch, A.~H. Toselli, E.~Vidal, Semiautomatic text baseline detection in
  large historical handwritten documents, in: 2014 14th International
  Conference on Frontiers in Handwriting Recognition, 2014, pp. 690--695.
\newblock \href {http://dx.doi.org/10.1109/ICFHR.2014.121}
  {\path{doi:10.1109/ICFHR.2014.121}}.

\bibitem{Moysset2015}
B.~Moysset, C.~Kermorvant, C.~Wolf, J.~Louradour, Paragraph text segmentation
  into lines with recurrent neural networks, in: 2015 13th International
  Conference on Document Analysis and Recognition (ICDAR), 2015, pp. 456--460.
\newblock \href {http://dx.doi.org/10.1109/ICDAR.2015.7333803}
  {\path{doi:10.1109/ICDAR.2015.7333803}}.

\bibitem{Pastor2016}
J.~Pastor-Pellicer, M.~Z. Afzal, M.~Liwicki, M.~J. Castro-Bleda, Complete
  system for text line extraction using convolutional neural networks and
  watershed transform, in: 2016 12th IAPR Workshop on Document Analysis Systems
  (DAS), 2016, pp. 30--35.
\newblock \href {http://dx.doi.org/10.1109/DAS.2016.58}
  {\path{doi:10.1109/DAS.2016.58}}.

\bibitem{Gruning2018}
T.~Gr{\"{u}}ning, G.~Leifert, T.~Strau{\ss}, R.~Labahn,
  \href{http://arxiv.org/abs/1802.03345}{{A Two-Stage Method for Text Line
  Detection in Historical Documents}}, CoRR.
\newline\urlprefix\url{http://arxiv.org/abs/1802.03345}

\bibitem{Ares2018}
S.~Ares~Oliveira, B.~Seguin, F.~Kaplan,
  \href{http://arxiv.org/abs/1804.10371}{dhsegment: {A} generic deep-learning
  approach for document segmentation}, CoRR abs/1804.10371.
\newblock \href {http://arxiv.org/abs/1804.10371} {\path{arXiv:1804.10371}}.
\newline\urlprefix\url{http://arxiv.org/abs/1804.10371}

\bibitem{Bukhari2012}
S.~S. Bukhari, T.~M. Breuel, A.~Asi, J.~El-Sana, {Layout analysis for Arabic
  historical document images using machine learning}, Proceedings -
  International Workshop on Frontiers in Handwriting Recognition (IWFHR) (2012)
  639--644\href {http://dx.doi.org/10.1109/ICFHR.2012.227}
  {\path{doi:10.1109/ICFHR.2012.227}}.

\bibitem{Baechler2011}
M.~Baechler, R.~Ingold, Multi resolution layout analysis of medieval
  manuscripts using dynamic mlp, in: 2011 International Conference on Document
  Analysis and Recognition, 2011, pp. 1185--1189.
\newblock \href {http://dx.doi.org/10.1109/ICDAR.2011.239}
  {\path{doi:10.1109/ICDAR.2011.239}}.

\bibitem{Wei2013}
H.~Wei, M.~Baechler, F.~Slimane, R.~Ingold, Evaluation of svm, mlp and gmm
  classifiers for layout analysis of historical documents, in: 2013 12th
  International Conference on Document Analysis and Recognition, 2013, pp.
  1220--1224.
\newblock \href {http://dx.doi.org/10.1109/ICDAR.2013.247}
  {\path{doi:10.1109/ICDAR.2013.247}}.

\bibitem{Cruz2012}
F.~C. Fernández, O.~R. Terrades, Document segmentation using relative location
  features, in: Proceedings of the 21st International Conference on Pattern
  Recognition (ICPR2012), 2012, pp. 1562--1565.

\bibitem{Lemaitre2008}
A.~Lemaitre, J.~Camillerapp, B.~Co{\"u}asnon,
  \href{https://doi.org/10.1007/s10032-008-0072-6}{Multiresolution cooperation
  makes easier document structure recognition}, International Journal of
  Document Analysis and Recognition (IJDAR) 11~(2) (2008) 97--109.
\newblock \href {http://dx.doi.org/10.1007/s10032-008-0072-6}
  {\path{doi:10.1007/s10032-008-0072-6}}.
\newline\urlprefix\url{https://doi.org/10.1007/s10032-008-0072-6}

\bibitem{Quiros2017}
L.~Quir{\'o}s, C.-D. Mart{\'i}nez-Hinarejos, A.~H. Toselli, E.~Vidal,
  Interactive layout detection, in: 8th Iberian Conference on Pattern
  Recognition and Image Analysis (IbPRIA), Springer International Publishing,
  Cham, 2017, pp. 161--168.

\bibitem{Zhong2015}
G.~Zhong, M.~Cheriet,
  \href{http://www.sciencedirect.com/science/article/pii/S0031320314003938}{Tensor
  representation learning based image patch analysis for text identification
  and recognition}, Pattern Recognition 48~(4) (2015) 1211 -- 1224.
\newblock \href
  {http://dx.doi.org/https://doi.org/10.1016/j.patcog.2014.09.025}
  {\path{doi:https://doi.org/10.1016/j.patcog.2014.09.025}}.
\newline\urlprefix\url{http://www.sciencedirect.com/science/article/pii/S0031320314003938}

\bibitem{Caruana93multitasklearning:}
R.~Caruana, Multitask learning: A knowledge-based source of inductive bias, in:
  Proceedings of the Tenth International Conference on Machine Learning, Morgan
  Kaufmann, 1993, pp. 41--48.

\bibitem{pix2pix2016}
P.~Isola, J.-Y. Zhu, T.~Zhou, A.~A. Efros, Image-to-image translation with
  conditional adversarial networks, arxiv.

\bibitem{SantosWZ17}
C.~N. dos Santos, K.~Wadhawan, B.~Zhou,
  \href{http://arxiv.org/abs/1707.02198}{Learning loss functions for
  semi-supervised learning via discriminative adversarial networks}, CoRR
  abs/1707.02198.
\newblock \href {http://arxiv.org/abs/1707.02198} {\path{arXiv:1707.02198}}.
\newline\urlprefix\url{http://arxiv.org/abs/1707.02198}

\bibitem{goodfellow2014generative}
I.~Goodfellow, J.~Pouget-Abadie, M.~Mirza, B.~Xu, D.~Warde-Farley, S.~Ozair,
  A.~Courville, Y.~Bengio,
  \href{http://papers.nips.cc/paper/5423-generative-adversarial-nets.pdf}{Generative
  adversarial nets}, in: Z.~Ghahramani, M.~Welling, C.~Cortes, N.~D. Lawrence,
  K.~Q. Weinberger (Eds.), Advances in Neural Information Processing Systems
  27, Curran Associates, Inc., 2014, pp. 2672--2680.
\newline\urlprefix\url{http://papers.nips.cc/paper/5423-generative-adversarial-nets.pdf}

\bibitem{suzuki1985topological}
S.~Suzuki, et~al., Topological structural analysis of digitized binary images
  by border following, Computer vision, graphics, and image processing 30~(1)
  (1985) 32--46.

\bibitem{perez1994}
J.-C. Perez, E.~Vidal, Optimum polygonal aprroximation of digitalized curves,
  Pattern Recognition Letters.

\bibitem{Quiros2018_b}
L.~Quir{\'o}s, L.~Serrano, V.~Bosch, A.~H. Toselli, E.~Vidal, From {HMMs} to
  {RNNs}: {C}omputer-assisted transcription of a handwritten notarial records
  collection, in: International Conference on Frontiers in Handwriting
  Recognition (ICFHR), 2018.

\bibitem{DBLP:RonnebergerFB15}
O.~Ronneberger, P.~Fischer, T.~Brox,
  \href{http://arxiv.org/abs/1505.04597}{U-net: Convolutional networks for
  biomedical image segmentation}, CoRR abs/1505.04597.
\newblock \href {http://arxiv.org/abs/1505.04597} {\path{arXiv:1505.04597}}.
\newline\urlprefix\url{http://arxiv.org/abs/1505.04597}

\bibitem{kingma2014adam}
D.~P. Kingma, J.~Ba, Adam: A method for stochastic optimization, 3rd
  International Conference on Learning Representations (ICLR).

\bibitem{paszke2016enet}
A.~Paszke, A.~Chaurasia, S.~Kim, E.~Culurciello, Enet: A deep neural network
  architecture for real-time semantic segmentation, arXiv preprint
  arXiv:1606.02147.

\bibitem{Simard03}
P.~Y. Simard, D.~Steinkraus, J.~Platt, Best practices for convolutional neural
  networks applied to visual document analysis, Institute of Electrical and
  Electronics Engineers, Inc., 2003.

\bibitem{Gruning2017}
T.~Gr{\"{u}}ning, R.~Labahn, M.~Diem, F.~Kleber, S.~Fiel,
  \href{http://arxiv.org/abs/1705.03311}{{READ-BAD:} {A} new dataset and
  evaluation scheme for baseline detection in archival documents}, CoRR
  abs/1705.03311.
\newblock \href {http://arxiv.org/abs/1705.03311} {\path{arXiv:1705.03311}}.
\newline\urlprefix\url{http://arxiv.org/abs/1705.03311}

\bibitem{long2015fully}
J.~Long, E.~Shelhamer, T.~Darrell, Fully convolutional networks for semantic
  segmentation, in: Proceedings of the IEEE conference on computer vision and
  pattern recognition, 2015, pp. 3431--3440.

\bibitem{Quiros2018}
L.~Quir{\'o}s, L.~Serrano, V.~Bosch, A.~H. Toselli, R.~Congost, E.~Saguer,
  E.~Vidal, \href{https://doi.org/10.5281/zenodo.1322666}{{Oficio de Hipotecas
  de Girona. A dataset of Spanish notarial deeds (18th Century) for Handwritten
  Text Recognition and Layout Analysis of historical documents.}} (Jul. 2018).
\newblock \href {http://dx.doi.org/10.5281/zenodo.1322666}
  {\path{doi:10.5281/zenodo.1322666}}.
\newline\urlprefix\url{https://doi.org/10.5281/zenodo.1322666}

\bibitem{diem2017}
M.~Diem, F.~Kleber, S.~Fiel, T.~Gruning, B.~Gatos,
  \href{doi.ieeecomputersociety.org/10.1109/ICDAR.2017.222}{cbad: Icdar2017
  competition on baseline detection}, in: 2017 14th IAPR International
  Conference on Document Analysis and Recognition (ICDAR), Vol.~01, 2017, pp.
  1355--1360.
\newblock \href {http://dx.doi.org/10.1109/ICDAR.2017.222}
  {\path{doi:10.1109/ICDAR.2017.222}}.
\newline\urlprefix\url{doi.ieeecomputersociety.org/10.1109/ICDAR.2017.222}

\bibitem{sanchez2016}
A.~Toselli, V.~Romero, M.~Villegas, E.~Vidal, J.~S{\'a}nchez,
  \href{https://doi.org/10.5281/zenodo.1297399}{Htr dataset icfhr 2016} (Feb.
  2018).
\newblock \href {http://dx.doi.org/10.5281/zenodo.1297399}
  {\path{doi:10.5281/zenodo.1297399}}.
\newline\urlprefix\url{https://doi.org/10.5281/zenodo.1297399}

\bibitem{Toselli2011}
A.~H. Toselli, E.~Vidal, F.~Casacuberta, Multimodal Interactive Pattern
  Recognition and Applications, Springer, Heidelberg, 2011.

\end{thebibliography}

\end{document}